\documentclass[letterpaper, 10 pt, conference]{ieeeconf}  

\IEEEoverridecommandlockouts                              

\usepackage{graphicx} 
\usepackage[hidelinks]{hyperref}
\usepackage{float}

\title{\LARGE \bf
DoUnseen: Tuning-Free Class-Adaptive Object Detection \\ of Unseen Objects for Robotic Grasping
}

\author{Anas Gouda and Moritz Roidl
\thanks{Both authors with TU Dortmund University, 44227 Dortmund, Germany
        {\tt\small firstname.lastname@tu-dortmund.de}}%
}

\begin{document}

\maketitle
\thispagestyle{empty}
\pagestyle{empty}

\begin{abstract}

How can we segment varying numbers of objects where each specific object represents its own separate class? To make the problem even more realistic, how can we add and delete classes on the fly without retraining or fine-tuning? This is the case of robotic applications where no datasets of the objects exist or application that includes thousands of objects (E.g., in logistics) where it is impossible to train a single model to learn all of the objects.

Most current research on object segmentation for robotic grasping focuses on class-level object segmentation (E.g., box, cup, bottle), closed sets (specific objects of a dataset; for example, YCB dataset), or deep learning-based template matching. In this work, we are interested in open sets where the number of classes is unknown, varying, and without pre-knowledge about the objects' types. We consider each specific object as its own separate class. 

Our goal is to develop an object detector that requires no fine-tuning and can add any object as a class just by capturing a few images of the object. Our main idea is to break the segmentation pipelines into two steps by combining unseen object segmentation networks cascaded by class-adaptive classifiers.

We evaluate our class-adaptive object detector on unseen datasets and compare it to a trained Mask R-CNN on those datasets. The results show that the performance varies from practical to unsuitable depending on the environment setup and the objects being handled.

The code is available in our DoUnseen library repository
\footnote{
\url{https://github.com/AnasIbrahim/image_agnostic_segmentation}
\label{footnote:github}
}.

\end{abstract}

\section{Introduction}

\begin{figure}
    \centering
    \includegraphics[width=0.48\textwidth]{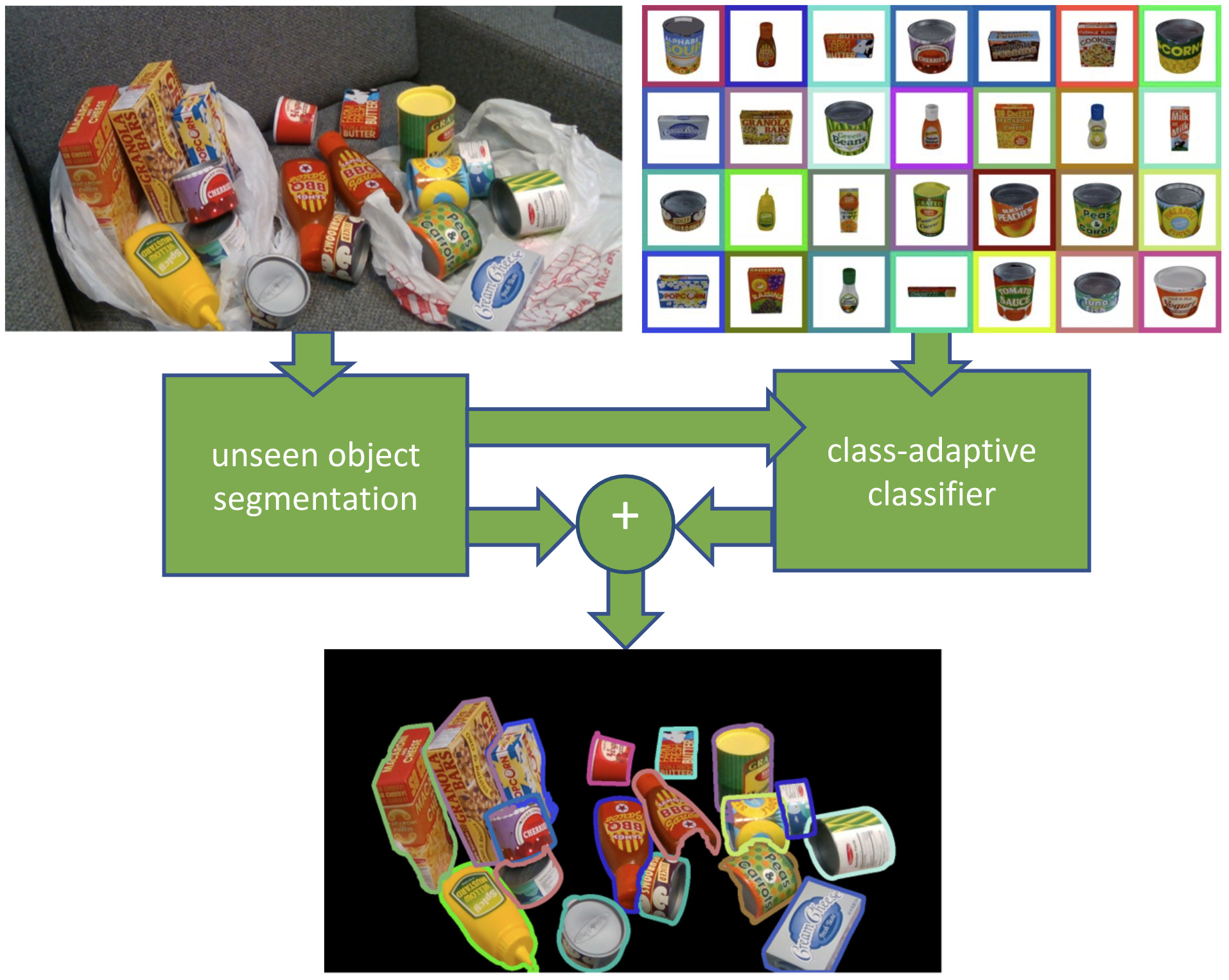}
    \caption{Our object detector building blocks. First stage, image is segmented into objects regardless of their class. Second stage, a class-adaptive classifier classifies the segmented masks into the best-matched object from the gallery set. The frame color of the objects in the segmented image represents the predicted class from the gallery set.}
    \label{fig:pipeline_teaser}
    \vspace{-0.6cm}
\end{figure}

Some robotic grasping applications do not require explicitly learning the object classes; they only require classification or re-identification from a set of images of the object (gallery set). This concept is referred to as template matching in classical non-learning based methods. The type of classes here differs from standard object detector classes (E.g., box, cup, bottle). In our case, each object would represent its own class, and object classes differ if objects are not identical (A whole-fat milk box and a low-fat milk box would be two different classes). The gallery set consists of a few images of each object covering all its unique faces. There are many cases where this concept would be the key to the solution.
The first case is applications in which collecting a dataset is not possible for reasons such as lack of technical expertise or time constraints.
Second, are the applications where the number of objects is large (in thousands). This is a real case that happens in logistics and industry. Even if a dataset of these thousands of objects exists, it would be challenging for an object detector to learn all the objects in a warehouse at once, and it may be impossible if many objects are highly similar. An object detector that can flexibly change the subset of classes used is a must.
Third, having a deep learning-based object detector without doing data collection and without training would be more convenient and easier.

The difference between deep learning-based template matching networks and class-adaptive object detectors is that the class-adaptive object detectors should be able to detect many object classes at once. In contrast, deep learning-based template matching aims to detect a single object class. Detecting a single object means linear time complexity for multiple objects; this is even more repelling as deep learning-based template matching networks are a sort of brute force. Also, class-adaptive object detectors should be aware of other objects in the environment that are not in the gallery set and cannot be classified.


Recent research in \cite{DTOID} \cite{MTTM} \cite{hu-2022} focused mainly on deep template matching. In this work, we are interested in going one step further than just deep template matching by developing a class-adaptive object detector for robotic grasping. The building blocks for our detector are shown in figure~\ref{fig:pipeline_teaser}. Unlike deep template matching methods that do the search in the whole embedding space, our method is more natural in the human sense as all objects are first segmented then the search is done per segmented mask.

\begin{figure*}
    \centering
    \includegraphics[width=0.9\textwidth]{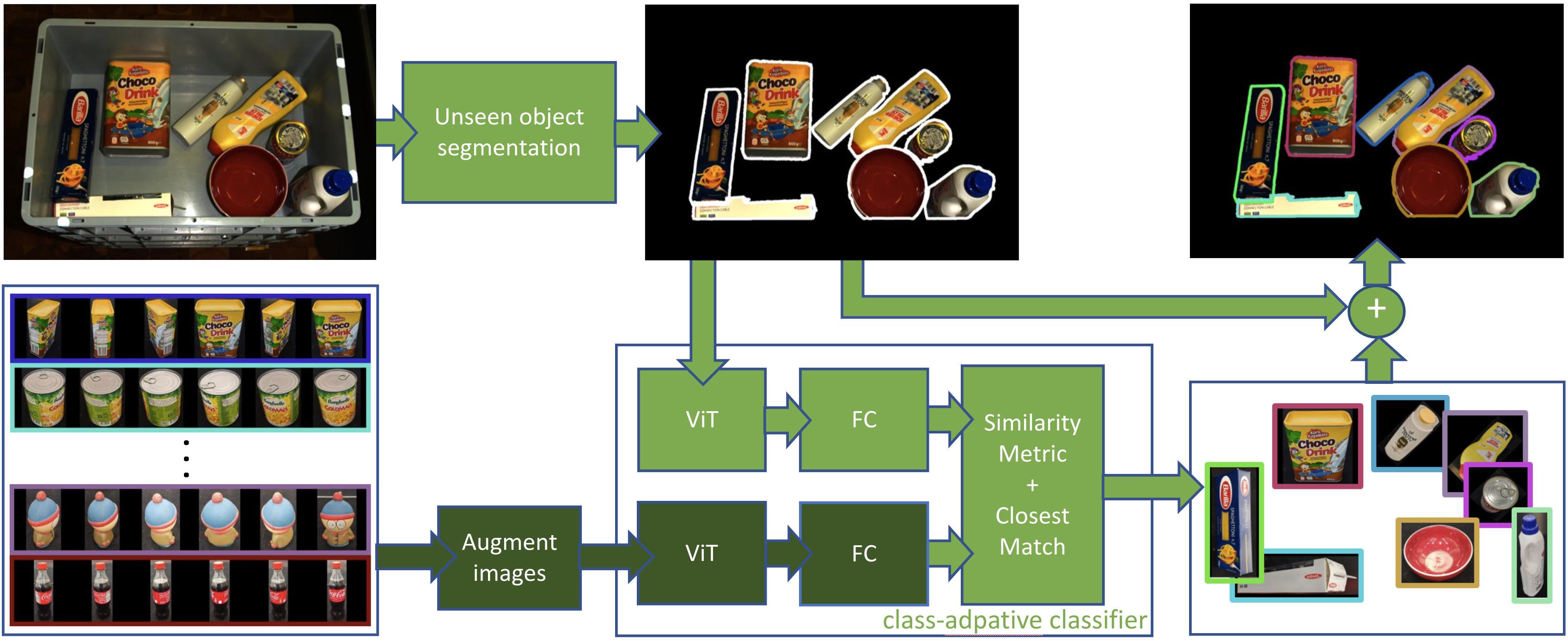}
    \caption{
    Our class-adaptive object detector. The unseen network segmentation is pre-trained in \cite{agnostic_seg}. The class-adaptive classifier is a siamese neural network. During testing, The gallery images are augmented, and their features are extracted and buffered and only recomputed when the gallery is changed (dark green blocks). This saves running time considerably for the siamese network.}
    \label{fig:detailed_method}
    \vspace{-0.6cm}
\end{figure*}
\section{Related Work}

There are different problems that work with unseen objects. Face detection, person re-identification, Deep template matching, unseen object instance segmentation and unseen 6D pose estimation are examples of such problems.

Several works study the problem of deep template matching in the context of robotic grasping.
DTOID \cite{DTOID} based on TDID \cite{TDID} introduced a network using RGB images for template matching network that uses a global template to specialize early features in the detection backbone and a local template that extracts view-point related features. Then a second stage is responsible for regressing the BBOX and the segmentation mask.
HU et al. \cite{hu-2022} used a similar network of DTOID \cite{DTOID} but enhanced the performance by constructing a NeRF model from the target object images for synthesizing more views of the target object.
MTTM \cite{MTTM} used RGB images for segmentation and pose estimation and used depth for ICP refinement.
These methods performed their evaluation mainly on the Linemod dataset \cite{linemod}. The Linemod dataset test set contains 15 objects and 15 scenes, with the goal is to detect one object in each scene. This case of Linemod is closer to template matching tests as the goal is to detect one object in the image. So, we opt out from testing on the Linemod dataset and use other datasets with multiple objects placed in different environments.


Related research for our object detector is based on two different research problems. The first problem is "unseen object instance segmentation" for robotic grasping, where the goal of the network is to segment all objects in the image without any knowledge about their type/class. The second related problem is class-adaptive classification.

The research on unseen object segmentation (also known as category-agnostic segmentation) is still an open problem, as it is hard to generalize between different environment setups. Different ideas were introduced for unseen object segmentation networks.
SD Mask R-CNN \cite{sdmaskrcnn} and Gouda et al. \cite{agnostic_seg} \cite{dopose} used depth images and RGB images consecutively to train variants of Mask R-CNN.
UCN \cite{UCN} used a CNN to compute embeddings followed by classical clustering.
UOIS \cite{UOIS} used a 2 stage segmentation method, the first stage used only depth to produce an initial mask and the second stage used RGB to produce the final mask.
MSMFormer \cite{MSMFormer} used a transformer architecture that simulates the von Mises-Fisher (vMF) mean shift clustering algorithm.
UOAIS \cite{UOAIS} went a step further by predicting the amodal segmentation mask (hidden parts of the object).
Instead of RGB or RGB-D, INSTR \cite{INSTR} used stereo image with a transformer architecture.




Several applications require classification/re-identification of unknown classes. Face detection \cite{facenet}, person re-identification \cite{osnet}, Vehicle re-identification \cite{veri}, Fashion classification \cite{DeepFashion}, and pallet re-identification \cite{palletblocks} are examples of such problems. Class-adaptive classification for robotic grasping is a similar problem to this category of problems.

Another work that presented a very close problem to our class-adaptive object detection is FewSol \cite{fewsol}. They created a dataset from existing datasets and newly collected data. FewSol benchmarked several few-shot object detectors and proposed the problem of joint object segmentation by combining unseen object segmentation and few-shot classification. The difference between our work and Fewsol is that it uses few-shot to learn new generic object classes (box, power drill, etc.), and we use the class-adaptive object detector to detect the gallery objects.



\begin{table}
    \centering
    \caption{Comparing different architecture as a backbone for the class-adaptive classifier on DoPose and HOPE datasets.}
    \resizebox{0.48\textwidth}{!}
    {
    \begin{tabular}{|c| c|c|c|c|c| c|c|c|c|c|}
        \hline
        backbone &  \multicolumn{5}{c|}{DoPose} & \multicolumn{5}{c|}{HOPE}\\
        \hline
            &  mAP & R1 & R5 & R10 & R20    & mAP & R1 & R5 & R10 & R20\\
        \hline
        \hline

        ResNet \cite{resnet} & 64.2 & 94.0 & 98.5 & 99.0 & 99.7 &   29.6 & 53.6 & 72.7 & 80.7 &  89.0 \\
        \hline
        DenseNet \cite{densenet} & 60.9 & 93.2 & 98.6 & 99.4 & \textbf{99.8}    & 28.6 & 59.3 & 76.5 & 82.5 & 88.8 \\
        \hline
        EfficientNet \cite{efficientnet} & 57.9 & 94.3 & 98.0 & 99.1 & 99.7   & 28.0 & 53.7 & 70.4 & 77.6 & 85.3\\
        \hline
        Wide-ResNet \cite{wideresnet} & 62.4 & 93.4 & 98.1 & 99.0 & 99.4    & 27.9 & 45.0 & 62.1 & 70.6 & 78.9 \\
        \hline

        ViT \cite{vit} & 56.5 & 91.5 & 97.5 & 98.7 & 99.4    & 28.6 & \textbf{62.6} & \textbf{79.2} & \textbf{87.1} & \textbf{91.4} \\
        \hline
        MaxVit \cite{maxvit} & 61.9 & 91.8 & 97.3 & 98.5 & 99.4    & 24.0 & 43.5 & 54.8 & 60.9 & 69.3 \\
        \hline
        SwinTransformer \cite{swintransformer} & \textbf{70.3} & 95.9 & \textbf{99.2} & \textbf{99.5} & 99.6    & \textbf{32.3} & 55.7 & 69.5 & 77.2 & 82.1 \\
        \hline

        OSNet-AIN \cite{osnetain} & 63.0 & \textbf{96.5} & 98.4 & 98.9 & 99.3    & 28.0 & 52.0 & 67.9 & 76.3 & 84.4 \\
        \hline

    \end{tabular}
    }
    \label{tab:backbones}
    \vspace{-0.6cm}
\end{table}

\begin{table}
    \caption{Validation of the class-adaptive classifier on the HOPE dataset.}
    \centering
        \begin{tabular}{|c| c|c|c|c|}
            \hline
                        & \multicolumn{4}{|c|}{HOPE}\\
            \hline
            Train Dataset & mAP & R1 & R5 & R10\\
            \hline
            \hline
            ImageNet  & 28.6 & 62.6 & 79.2 & 87.1 \\
            \hline
            FewSol    & 38.8 & 56.0 & 73.9 & 81.5 \\
            \hline
        \end{tabular}
    \label{tab:classifier_training}
    \vspace{-0.6cm}
\end{table}

\begin{table*}
    \caption{
    Evaluation of our object detector against a trained method on DoPose and HOPE datasets using COCO metrics.
    }
    \resizebox{0.999\textwidth}{!}
    {
    \centering
    \begin{tabular}{|c|    c|c|c|c|c|c|c|c|    c|c|c|c|c|c|c|c|}
        \hline
                   & \multicolumn{8}{|c|}{DoPose} & \multicolumn{8}{|c|}{HOPE} \\
        \hline
                   & \multicolumn{4}{|c|}{BBox} & \multicolumn{4}{|c|}{Seg} & \multicolumn{4}{|c|}{BBox} & \multicolumn{4}{|c|}{Seg} \\
        \hline
                   & mAP & AP50 & AP75 & AR    & mAP & AP50 & AP75 & AR    & mAP & AP50 & AP75 & AR    & mAP & AP50 & AP75 & AR\\
        \hline
        \hline
        Mask R-CNN                                    & 84.5 & 99.3 & 96.5 & 88.3 & 74.7 & 97.1 & 86.4 & 79.0        & 23.8 & 34.9 & 27.9 & 35.5 &  9.5 & 17.7 & 10.0 & 15.7\\
        \hline
        GT mask + our class.                          & 79.0 & 79.0 & 79.0 & 86.0 & 79.0 & 79.0 & 79.0 & 86.0        & 47.1 & 47.1 & 47.1 & 62.3 & 47.1 & 47.1 & 47.1 & 62.3        \\
        \hline
        Unseen seg. \cite{agnostic_seg} + our class.  & 51.0 & 69.5 & 61.2 & 62.0 & 45.1 & 68.1 & 52.9 & 55.3   & 28.4 & 40.4 & 33.8 & 37.3 & 30.3 & 40.3 & 31.3 & 39.5\\
        \hline
    \end{tabular}
    }


    \label{tab:eval_trained_methods}
    \vspace{-0.6cm}
\end{table*}

\section{Method}

As pre-mentioned, there are two methods to handle unseen objects in robotic grasping; First Deep template matching and second class-adaptive object detection. Other than the benefit of the class-adaptive classifier of simultaneously detecting multiple objects. More benefits include allowing the two building blocks (segmentation and classification) to be developed separately, giving wider space for explainability, and different methods to be tested on each problem. Moreover, the second method can easily adapt solutions from related problems (E.g., person re-identification, and face detection) and allow using refinement methods on each block (E.g., merging and splitting of over/under-segmented objects \cite{rice}).

Our class-adaptive object detector consists of two stages. For the first stage (unseen object segmentation), we use a previous work from \cite{agnostic_seg}. Other more sophisticated methods \cite{UCN} \cite{UOIS} \cite{MSMFormer} for unseen object segmentation exists, however they don't generalize well on complex test environment. While these methods still could be fine-tuned/tweaked for the various test environments, this is outside the scope of this work. For the second stage (the class-adaptive classifier), we develop a siamese neural network. Both the segmentation and classification methods are RGB only. The training of both the segmentation network and the the classifier cannot include any test objects and cannot use the test objects for fine-tuning.

Figure~\ref{fig:detailed_method} shows our class-adaptive object detector in detail. The detector first segments all the objects in the image without any knowledge about their classes. Second, the detector classify/match these segmented patches to one of the objects in the gallery set. This way, a dataset is not required; only a few images of each object are enough. Changing the classes would be possible by just changing the gallery set. Handling thousands of objects from a big gallery would be possible by using a candidate subset only from that big gallery for the ongoing situation. For example, a big gallery would be all objects in a warehouse, and the candidate subset would be the list of objects stored in a bin during bin picking. This big gallery of object images typically exist for most warehouses and retailers.


The image to be segmented can include any number of objects with any number of instances per object. The segmentation method must differentiate between the different instances of the same object. The gallery set is not limited to the objects present in the scene and can include other objects not presented (in our experiments we include all objects from the test datasets). Each object in the gallery set should include N number of images of each object.

\subsection{Training and Evaluation Datasets}

As our object detector consists of two modules, we use different datasets to train each module. We test the entire object detector on datasets different than the ones used for the training. The unseen object segmentation network in \cite{agnostic_seg} that we use for our object detector is trained on NVIDIA FAT synthetic dataset \cite{FATdataset} only. For training our class-adaptive classifier, we require a dataset with many objects, with each object captured from many perspectives. This dataset needs to be parsed first to isolate objects to suit the expected input of the classifier. Each object in each image is cropped around its BBox then superimposed by its binary segmentation mask. Then all patches of each object from all images are combined as a class regardless of their original image.

There are two possibilities for datasets to train the classifier. The first possibility is the datasets from the related problem of 6D Pose estimation. DoPose \cite{dopose}, HOPE \cite{tyree2022hope}, Linemod \cite{linemod}, T-Less \cite{tless}, HomeBrew \cite{homebrew},  YCB-V \cite{ycb} are examples of such datasets. Each dataset contains 18, 28, 15, 30, 33, and 21 objects, respectively. The second possibility is the FewSol dataset \cite{fewsol} for few-shot learning. The portion of FewSol suitable for our problem includes 666 objects (336 real objects and 330 synthetic objects from Meta dataset\cite{googlemetadataset}). The number of occurrences of objects in the pre-mentioned 6D Pose estimation datasets is low (18-33), but the number of occurrences per object is high (hundreds to thousands; as per comparison in \cite{dopose}). On the other hand, the number of objects in FewSol is high (666), but the number of occurrences is low (only 9 images per object). The ideal dataset would contain a large number of objects with a large number of occurrences in different conditions and poses. We use the FewSol dataset to train our classifier as the number of objects is more crucial to the training, and we depend on data augmentation to introduce more occurrences per object.


For the validation of our classifier, we use the DoPose and the HOPE datasets. The first reason for this choice is that DoPose and HOPE datasets scenes are captured in several environments and record a few images per each unique scene giving more variance in the testing data. Unlike other datasets (Linemod, T-Less, HomeBrew, YCB, TYO-L), which are captured in video frames generating hundreds of frames per each unique scene. The second reason is that both datasets represent the hardest and easiest cases. Figure~\ref{fig:test_datasets} plots the objects from both dataset on the dominant color axis and shape axis. The DoPose dataset objects have distinctive shapes with distinctive colors. In contrast, many object from the HOPE dataset exhibit much higher similarity or identity for both shape and color. This makes the DoPose dataset easy to differentiate and the HOPE dataset more challenging. Our classifier requires a gallery set and query set. The gallery set consists of 2-10 real images per object in isolation covering all its unique faces. The gallery images are augmented by rotating multiples of 45 degrees, generating 8X more images. The query set consists of cropping each occurrence of each object around its BBOX and superimposing its binary mask. Then all occurrences of each object represent a class in the query set.

For testing the whole object detector, we also use the DoPose and the HOPE datasets.

\subsection{The Classifier Architecture and Evaluation Metrics}

The class-adaptive classifier is a siamese neural network, giving a score for the similarity between 2 images (one query image and one gallery image). For validation of the classifier, we use the Cumulative Matching Characteristics (CMC) metrics which are used for similar re-identification/classification problems. We use the evaluator from the Torchreid library \cite{torchreid}. The CMC metrics ranking (R1:R20) represent the accuracy of the top-k samples being correctly matched from the gallery images for each query image.

But which backbone is more suitable for our problem? Table~\ref{tab:backbones} shows the evaluation of different backbones pre-trained on ImageNet on the validation sets of the DoPose and the HOPE datasets. This evaluation is carried out using the Torchreid library. We use the biggest instance of all backbone architectures (ResNet-152, etc.). As expected, all backbones score higher on the DoPose dataset than on the HOPE dataset. For the DoPose dataset, SwinTransformer scores highest on mAP, and OSNET-AIN scores the highest rank-1 even if the OSNet uses only 2.3 million parameters, considerably less than all other architectures. For the HOPE dataset, SwinTransformer scores highest for mAP, and ViT scores the highest rank-1. Classification of DoPose objects looks like an easy challenge for most backbones, while HOPE is not. So the backbone we choose for our classifier is ViT, as it is the most promising in classifying the more challenging case of the HOPE dataset.

For our Siamese network, we use the ViT backbone 'vit\_b\_16' instance from the PyTorch library. There is a fully connected layer cascaded after each backbone with the same output size as the input size (768). Then the output from both FC layers is passed to the cosine similarity as shown in figure~\ref{fig:detailed_method}.

We carry out an evaluation of the whole object detector (segmentation + classification). For this evaluation, we use the COCO metrics, which will allow us to compare our class-adaptive object detector with a standard trainable object detector.


\section{Evaluation}

\subsection{Class Adaptive Classifier Training}



The training runs for 10 epochs on the FewSol dataset, with the ViT backbone frozen for the first epoch. The training data is sampled to feed 50\% a query and a gallery image from the same class, with the output being set to one and 50\% from different classes, with the output being zero. Table~\ref{tab:classifier_training} shows the validation of the training compared to the pretraining on ImageNet. We notice that the mean average precision increased by 10.2\%, but the rank-1 decreased by 6.6\%. This means that the model gives a better score between gallery and query images but a lower score for the closest image. This shows that our network is effective, but there is a need for a dataset with a large number of objects with a large number of images (occurrences) per object. Achieving a higher mAP could be made by depending on the centroid of the gallery image in the feature space rather than the closest image.



\subsection{Evaluation against Trained Methods}






How would our class-adaptive object detector perform against a model trained on the test objects? We train a Mask R-CNN model using the DoPose and the HOPE datasets. Table~\ref{tab:eval_trained_methods} shows the comparison using COCO metrics. We first conduct the evaluation with the GT masks (Ground Truth) and our class-adaptive classifier. Then Second, with the unseen segmentation method from \cite{agnostic_seg} combined with our class-adaptive classifier.

For the DoPose dataset, the performance drops from the trained method from 84.5\% to 79.0\% when using GT masks and our class-adaptive classifier and drops from 84.5\% to 51.0\% when using both unseen object segmentation from \cite{agnostic_seg} and our class-adaptive classifier. This shows that our classification is comparable to trained methods when segmented masks are accurate and objects are easy to distinguish. For the HOPE dataset, we can see that the performance of our untrained method is higher than the trained methods. An important note here is that the images of the HOPE-Image dataset are blurry, and this reduces the accuracy of the trained methods (Mask R-CNN). Even with the GT masks the class-adaptive classification performance is still low.


\section{Conclusion and Future Work}






In this work, we illustrated how a class-adaptive object detector for robotic grasping is realized. Our detector could be practical to use if the dataset is not highly cluttered and objects are easily distinguishable. Increasing the performance of class-adaptive object detectors can evolve in two ways. First, for the segmentation methods to generalize on environment setups. Second, for the classification methods to be able to handle a higher degree of similarity. A big hurdle that faces the development of the class-adaptive classifier is the datasets. Either large (thousands) high-quality CAD models need to be collected to produce a synthetic dataset, or a huge real dataset needs to be collected. Such huge dataset collection was recently published in the ArmBench dataset \cite{mitash2023armbench}.


\clearpage

\section{Acknowledgement}
This work is funded by the German Federal Ministry of Education and Research (BMBF) in the course of the the Lamarr Institute for Machine Learning and Artificial Intelligence under grant number LAMARR23B.


\bibliographystyle{IEEEtran}
\bibliography{IEEEabrv,./references.bib}

\onecolumn

\appendix

\begin{figure}[H]
    \centering
    \includegraphics[width=0.4\textwidth]{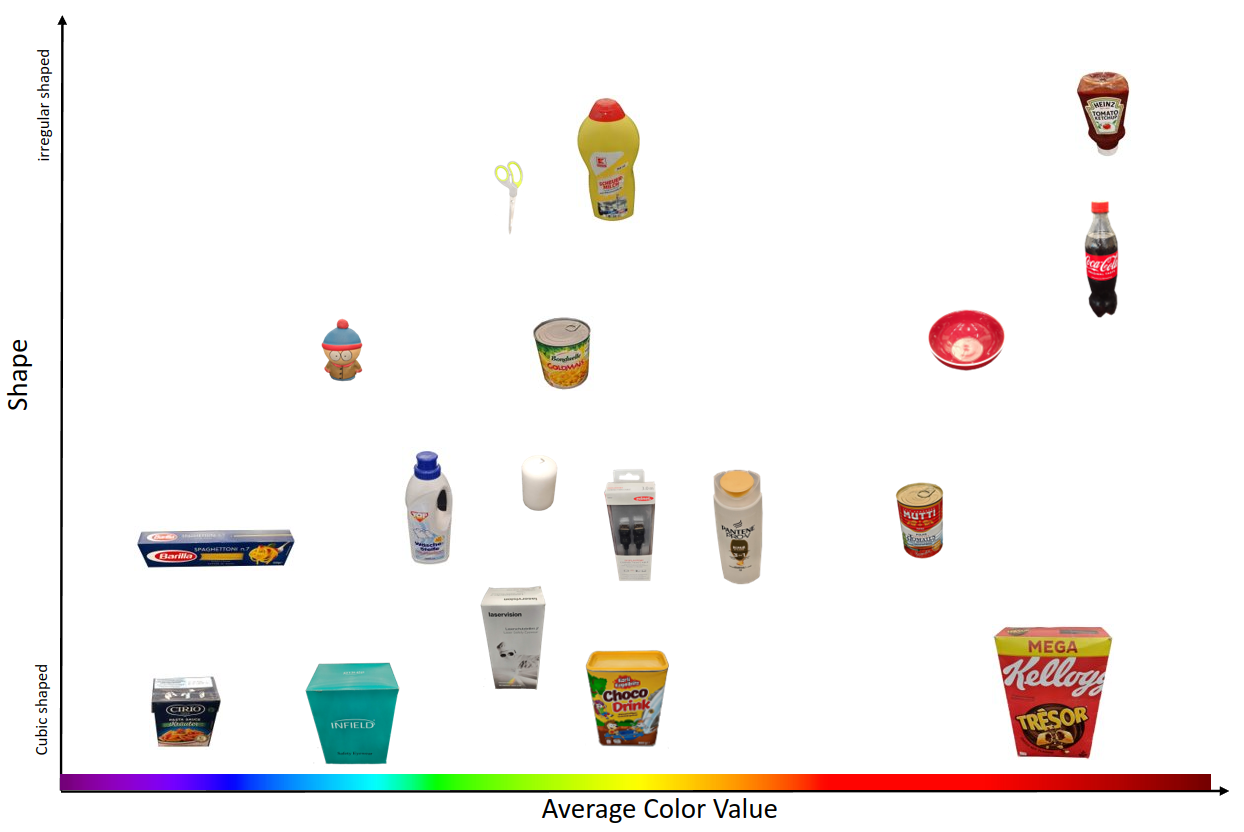}
    \includegraphics[width=0.4\textwidth]{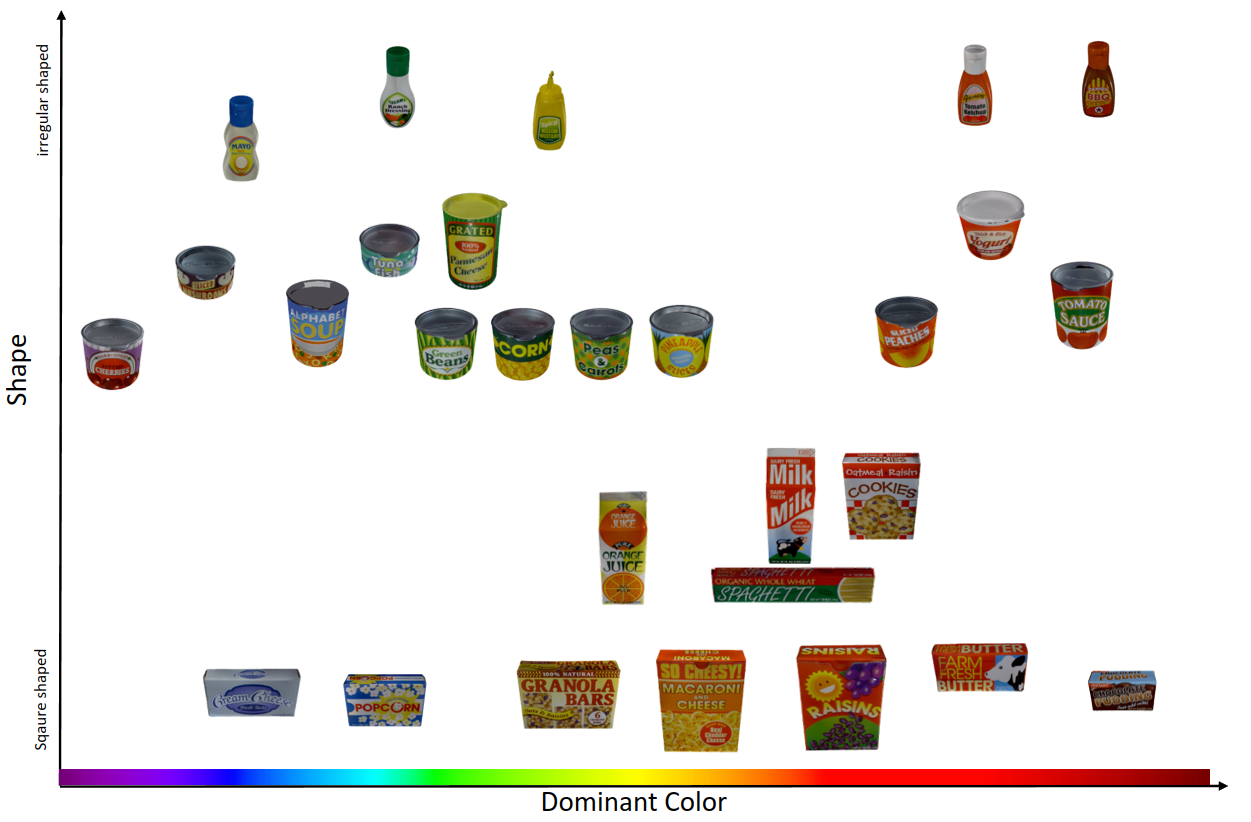}
    \caption{Testing dataset: DoPose (left) and HOPE (right). The two datasets exhibit different difficulty levels. DoPose includes 18 objects that are easily distinguished as the color and shape of the objects are quite different. HOPE dataset includes 28 objects that are hard to distinguish as the color and shape of many objects are similar or identical. These datasets will enable us to study the extremes, the easier case of DoPose, and the more challenging case of HOPE.}
    \label{fig:test_datasets}
    \vspace{-0.6cm}
\end{figure}

\end{document}